\documentclass[letterpaper, 10 pt, conference]{ieeeconf}
\IEEEoverridecommandlockouts                              % This command is only needed if
\usepackage[T1]{fontenc}
                               
\usepackage{graphics} 

\usepackage{amsmath} 
\usepackage{amssymb} 
\usepackage{graphicx}
\usepackage{glossaries}
\usepackage{multirow}
\usepackage{subcaption}
\usepackage{array}
\usepackage{booktabs}
\usepackage{caption}
\usepackage{xcolor}
\usepackage{diagbox}
\usepackage{slashbox}
\usepackage{hyperref}
\hypersetup{
    colorlinks=true,
    linkcolor=blue,
    filecolor=magenta,      
    urlcolor=cyan,
    pdftitle={Overleaf Example},
    pdfpagemode=FullScreen,
    }

\def\secref#1{Sec.~\ref{#1}}
\def\figref#1{Fig.~\ref{#1}}
\def\tabref#1{Tab.~\ref{#1}}
\def\eqref#1{Eq.~(\ref{#1})}

\usepackage{amsopn}

\newcommand{\bR}{\mathbf{R}}

\newcommand{\bT}{\mathbf{T}}

\newcommand{\bc}{\mathbf{c}}
\newcommand{\be}{\mathbf{e}}

\newcommand{\bm}{\mathbf{m}}

\newcommand{\bx}{\mathbf{x}}

\newcommand{\bz}{\mathbf{z}}

\newcommand{\bn}{\mathbf{n}}

\newcommand{\bp}{\mathbf{p}}

\DeclareMathOperator*{\argmin}{argmin}

\def\secref#1{Sec.~\ref{#1}}
\def\figref#1{Fig.~\ref{#1}}
\def\tabref#1{Tab.~\ref{#1}}
\def\eqref#1{Eq.~(\ref{#1})}

\makeatletter
\DeclareRobustCommand\onedot{\futurelet\@let@token\@onedot}
\def\@onedot{\ifx\@let@token.\else.\null\fi\xspace}

\define@key{FV}{highlightlines}{\edef\FV@HighlightLinesList{#1}}

\makeatother
\def\g2o{$g^2o$}

\newcommand{\RR}{\mathbb{R}}

\def\argmin{\mathop{\rm argmin}}
\newacronym{slam}{SLAM}{Simultaneous Localization and Mapping}
\newacronym{sfm}{SfM}{Structure from Motion}
\newacronym{gn}{GN}{Gauss-Newton}
\newacronym{ils}{ILS}{Iterative Least-Squares}
\newacronym{icp}{ICP}{Iterative Closest Point}
\newacronym{lm}{LM}{Levenberg-Marquardt}
\newacronym{gf}{GF}{Gaussian Filters}
\newacronym{pf}{PF}{Particle Filters}
\newacronym{sdp}{SDP}{Semi-Definite Programming}
\newacronym{bst}{BST}{Binary Search Tree}
\newacronym{ndt}{NDT}{Normal Distributed Transform}
\newacronym{ba}{BA}{Bundle Adjustement}
\newacronym{fmcw}{FMCW}{Frequency Modulated Continuous Wave}
\newacronym{imu}{IMU}{Inertial Measurement Unit}
\newacronym{lidar}{LiDAR}{Light Detection and Ranging}
\newacronym{tof}{ToF}{Time of Flight}
\newacronym{mems}{MEMS}{Micro-electromechanical Systems}
\newacronym{iekf}{IEKF}{Iterative Extended Kalman Filter}
\newacronym{std}{STD}{Standard Deviation}
\newacronym{irls}{IRLS}{Iterative Reweighted Least-Squares}
\newacronym{rmse}{RMSE}{Root Mean Square Error}

\title{\LARGE \bf Enhancing LiDAR performance: Robust De-skewing Exclusively Relying on Range Measurements}
\author{Omar Salem \and Emanuele Giacomini \and Leonardo Brizi \and Luca Di Giammarino \and Giorgio Grisetti% <-this % stops a space
  \thanks{All authors are with the Department of Computer, Control and Management Engineering, La Sapienza University of Rome, Rome 00185, Italy}
  \thanks{This work has been supported by PNRR MUR project PE0000013-FAIR}
  }
\begin{document}

% \authorrunning{O. Salem et al.}

% \institute{$^{1}$Dipartimento di Ingegneria informatica, automatica e gestionale
% \\Sapienza University of Rome, Rome, Italy \\\email{lastname@diag.uniroma1.com}}

%%%%%%%%%%%%%%%%%%%%%%%%%%%%%%%%%%%%%%%%%%%%%%%%%%%%%%%%%%%%%%%%%%%%%%%%%%%%%%%%
\maketitle
\thispagestyle{empty}
\pagestyle{empty}
\begin{abstract}
  
  Most commercially available Light Detection and Ranging
(LiDAR)s measure the distances along a 2D section of the environment
by sequentially sampling the free range along directions centered at the
sensor's origin. When the sensor moves during the acquisition, the measured ranges are affected by a phenomenon known as "skewing", which
appears as a distortion in the acquired scan. Skewing potentially affects
all systems that rely on LiDAR data, however, it could be compensated
if the position of the sensor were known each time a single range is measured. Most methods to de-skew a LiDAR are based on external sensors
such as IMU or wheel odometry, to estimate these intermediate LiDAR
positions. In this paper, we present a method that relies exclusively on
range measurements to effectively estimate the robot velocities which are
then used for de-skewing. Our approach is suitable for low-frequency LiDAR where the skewing is more evident. It can be seamlessly integrated
into existing pipelines, enhancing their performance at a negligible computational cost.
  
  \end{abstract}
  
  % \footnotetext[]{}
  %%%%%%%%%%%%%%%%%%%%%%%%%%%%%%%%%%%%%%%%%%%%%%%%%%%%%%%%%%%%%%%%%%%%%%%%%%%%%%%%
  \section{Introduction}
  \label{sec:intro}
  
  Accurate and reliable mapping, localization, and navigation are essential for a
  wide range of robotics applications from autonomous driving, logistics, search
  and rescue, and many others. To this extent, \gls{lidar} sensors are a popular choice
  since they allow us to sense both the free space and the location of obstacles around
  the robot. A planar \gls{lidar} measures the distance of an object by deflecting laser
  beams around the sensor's axis of rotation.

      \begin{figure*}[hbt]
        \centering
        \begin{subfigure}{0.45\textwidth}
            \centering
            \includegraphics[width=\textwidth]{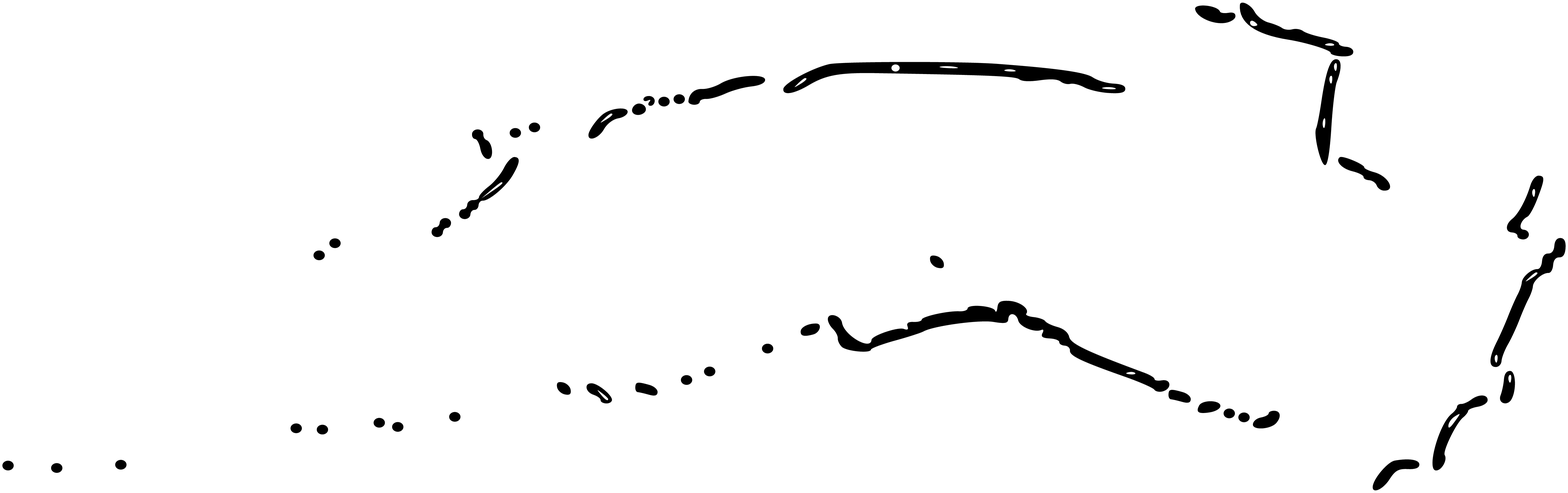}
            \subcaption{}
            \label{fig:skewed_scan}
        \end{subfigure}
        \hfill
        \begin{subfigure}{0.45\textwidth}
            \centering
            \includegraphics[width=\textwidth]{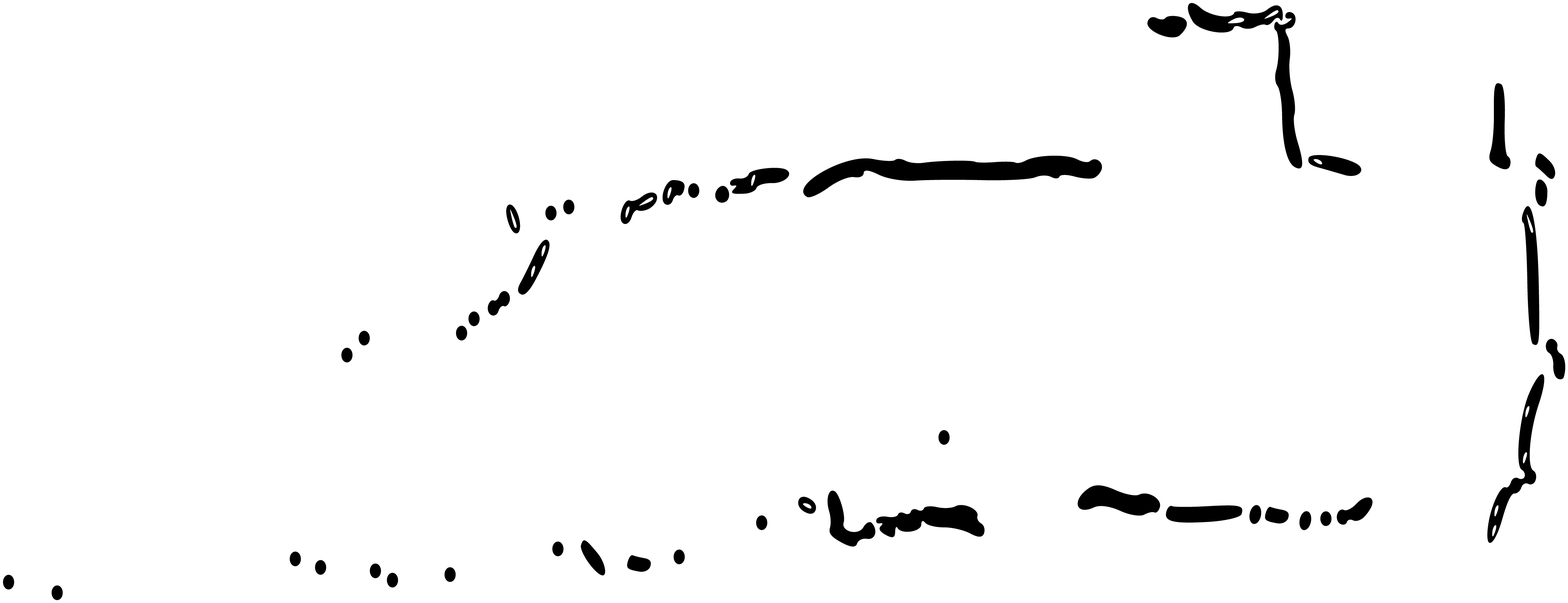}
            \subcaption{}
            \label{fig:deskewed_scan}
        \end{subfigure}
        \begin{subfigure}{0.45\textwidth}
          \centering
          \includegraphics[width=\textwidth]{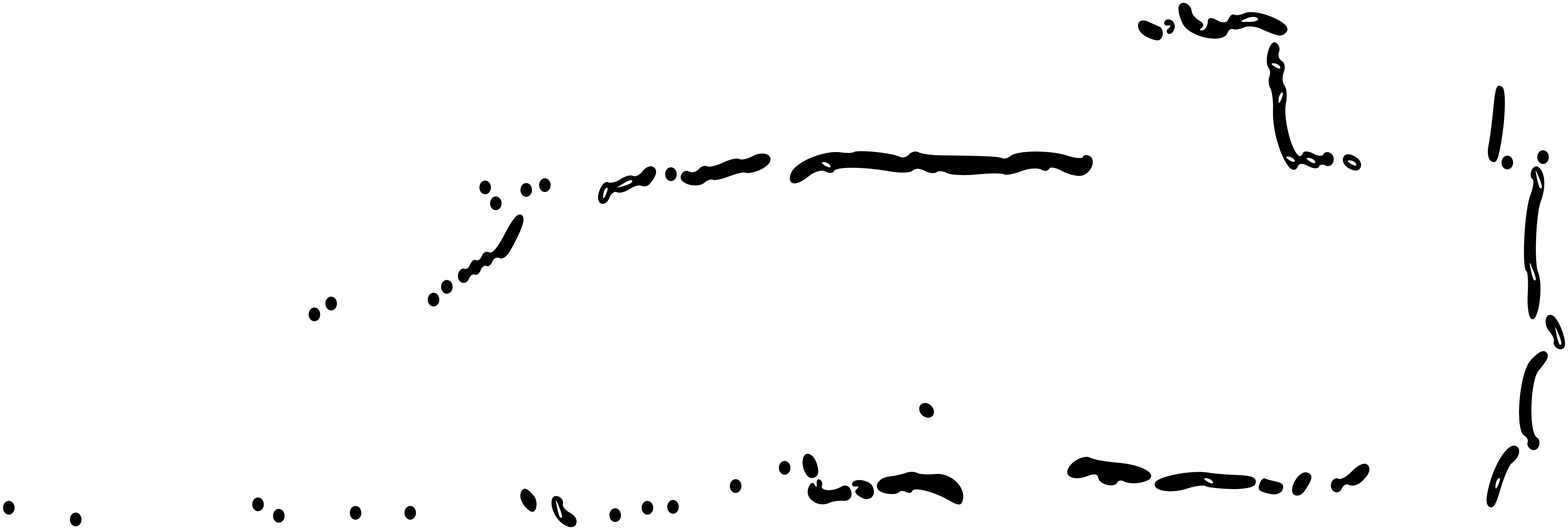}
          \subcaption{}
          \label{fig:gt_scan}
      \end{subfigure}
  \caption{Effect of de-skewing. (a) Unprocessed (raw) scan acquired by a rotating robot, while (b) is the corresponding ground truth scan, and finally (c) is the same scan in (a), processed (de-skewed) using our proposed approach. }
  \label{fig:motivation}
  \end{figure*}

  In this work we focus on low-grade 2D \gls{lidar}s such as the
  \href{https://www.slamtec.com/en/Lidar}{InnoMaker-LD-06} or the \href{https://www.slamtec.com/en/Lidar}{RPI-Lidar A1}
  which at the time of writing can be bought for less
than 100 Eur. The major shortcoming of these devices is their relatively slow
angular speed which results in a full sweep of measurements becoming available
at a frequency between 5 to 10 Hz. In contrast to more expensive models, these devices are not equipped with an inertial measurement unit or other means to
estimate the proprioceptive motion of the sensor while it moves.
In such devices, the scan acquisition(a full sweep of laser beams around the
vertical axis of the sensor) is not instantaneous, hence when the robot carrying
the LiDAR moves, the origin of the beam in the world frame changes for each
sensed range. Neglecting this fact, and assuming that all measurements gathered
during one rotation are sampled at the same time resulting in a distorted or skewed
scan as shown in \figref{fig:motivation}. Despite this consideration, most perception subsystems
in a navigation stack treat the scans as rigid bodies, however, the effect of skewing
leads to undesirable decay in accuracy that can result in failures as analyzed by Al-Nuaimi~\emph{et al.}~\cite{al2016analyzing}.
  
A common way to compensate for the motion of the \gls{lidar}
is to integrate external proprioceptive measurements (dead
reckoning and/or \gls{imu}), to estimate the origin and orientation
of the laser beam each time a range is
acquired~\cite{he2020skewing,DBLP:conf/rss/ZhangS14,legoloam2018}.
High-end 3D \gls{lidar}s already contain a synchronized \gls{imu},
which is not available on the inexpensive models previously mentioned.

In this paper, we propose an approach to estimate the velocities of the robot carrying the \gls{lidar}, based \emph{solely} on the range
measurements, that are processed as a stream. Our method is based on
a non-rigid plane-to-plane registration algorithm, where the velocity
of the platform carrying the sensor is estimated by maximizing the geometric consistency
of the stream of ranges. We conducted several statistical experiments on
synthetic and real data. Results verify that our de-skewing method is effective
in estimating the motion of the robot, at negligible computation.

 In \figref{fig:motivation} we illustrate the effects of
our approach on measurements acquired with a constantly rotating \gls{lidar}, with a rotational velocity
3 rad/s. In the remainder of this paper, we first
review the related work in \secref{sec:related}, subsequently we
describe in detail our de-skewing method for 2D \gls{lidar}
data \secref{sec:main}. We conclude the paper by presenting some synthetic
and real results in \secref{sec:exp}, that show the ability of the proposed system. And finally, we
draw some conclusions on the benefits and the limitations of our
approach.

  %%%%%%%%%%%%%%%%%%%%%%%%%%%%%%%%%%%%%%%%%%%%%%%%%%%%%%%%%%%%%%%%%%%%%%%%%%%%%%%%
  
  \section{Related Work}
  \label{sec:related}
  
  In this section, we review the most recent approaches to de-skew
  \gls{lidar} data. 
  Existing approaches for de-skewing mostly use either wheeled odometry
  or \gls{imu} to estimate the relative motion of the sensor/robot while a scan is being acquired. This process involves motion integration relying on estimators that process the raw proprioceptive
  data such as a filter or an integrator that directly provides an
  estimate of the sensor position each time a single range is measured.

  These methods fall in the class of \emph{loosely-coupled}, since they
  do not use the \gls{lidar} information to refine the proprioceptive
  estimate. Among this class, we find the work of Tang~\emph{et
    al.}~\cite{tang2015lidar} that proposes to match subsequent scans to
  compute the relative motion, where the initial relative orientations are
  provided by an \gls{imu}. 
  Subsequently, He~\emph{et al.}~\cite{he2020skewing} proposed a method
  to estimate relative motion between \gls{imu} poses and de-skew
  subsequent ranges by using these smoothed poses.
   
  In contrast to loosely-coupled approaches, \emph{tightly-coupled} methods
  jointly process \gls{lidar} and \gls{imu}. These methods are typically
  based on either smoothing or filtering. Wei ~\emph{et
    al.}~\cite{xu2021fast} uses a pre-integration scheme to estimate the
  \gls{lidar}'s ego-motion based on the inertial measurements, and
  updates the IMU biases in the correct step of an iterative EKF
  whenever a scan is completed. Shan ~\emph{et al.} \cite{shan2020lio} includes the states
  of \gls{imu} in the factor-graph, updating \gls{imu} biases through
  the optimization process, adapting a well-known computer vision work
  \cite{forster2016manifold} to the \gls{lidar} case. These ideas have
  been further extended in \cite{wang2022d,ye2019tightly}
  where the authors resort to full factor-graph optimization to obtain a
  state configuration that is maximally consistent with all the
  \gls{imu} measurements and \gls{lidar} ranges.

  For coupled approaches, accurate time synchronization between the sensors is essential.
    Unfortunately, this is not straightforward to obtain on inexpensive small
  devices due to the unpredicted communication latencies that affect the communication channels. 
  These issues, however, can be completely avoided when using only \gls{lidar} data to perform
  de-skewing. At their core \gls{lidar} only methods have a
  registration algorithm, which aims to compute the velocities of the
  sensor while the robot moves. The basic intuition is that if the
  correct velocities were found, the sequence of range measurements
  could be assembled in a maximally consistent scan. In this context, 
  Moosman \emph{et al.}~\cite{moosmann2011velodyne} propose to linearly interpolate
  scans between the last known pose, and the currently estimated pose, de-skewing the scan at both poses. 
  However, this double approximation might hinder the accuracy when the
  initial registration fails, resulting in even worse estimates.

  Al-Nuaimi \emph{et al.}~\cite{al2016analyzing} propose a weighting
  schema based on their adapted Geometric Algebra LMS solver, where they
  assume that consecutive relative translation and angular motion are equal,
   which might be inaccurate due to the existence of drift and friction.

  In this paper, we propose a planar \gls{lidar}-only approach that
  addresses the de-skewing problem by continuously registering the
  the sequence of range measurements. The registration is carried on by
  estimating the \gls{lidar}'s planar velocity, which iteratively
  minimizes a plane-to-plane metric between the de-skewed endpoints. Our
  results demonstrate that our methodology is accurate in estimating 
  the velocity of the robot, only based on the laser
  measurements.

  %%%%%%%%%%%%%%%%%%%%%%%%%%%%%%%%%%%%%%%%%%%%%%%%%%%%%%%%%%%%%%%%%%%%%%%%%%%%%%%%
  \section{Our Approach}
  \label{sec:main}
  
  Our method leverages some mild assumptions about the motion of the sensor mounted on the robot and
  the structure of the environment to operate on a continuous stream
  of range measurements. More specifically we assume to have a
  2D slow \gls{lidar} sensor, hence,  we define each measurement as
  $\bz_i=\left<r_i, \alpha_i, t_i \right>$. Here, $r_i$ is the
  range, $\alpha_i$ is the angle of the laser beam with respect to the origin of
  the sensor and $t_i$ is the timestamp. We assume the environment
  consists of a locally smooth surface and that the robot velocities change mildly within a \gls{lidar} beam revolution.
  
  Our method estimates these velocities by
  registering the stream of measurements $\{\bz_i\}$ onto itself. The
  most likely velocities are the ones that if applied for de-skewing
  renders the measurement maximally consistent. Consistency is measured
  by a plane-to-plane metric applied between the corresponding \gls{lidar} endpoints. More formally, let $\bx=\left(v \hspace{0.2 cm} \omega
  \right)^T$ be the translational and angular velocity of the robot we want
  to estimate $\bx^*$, such that
  \begin{equation}
    \bx^* = \argmin_{\bx} \sum_{\left<i,j\right>\in\mathcal{C}} 
    \rho \lVert \be(\bx,t_i,t_j)\rVert^2.
    \label{eq:cost}
  \end{equation}
  Where $\be(\bx,t_i,t_j)$ denotes an error vector between two planar
  scan patches computed around two corresponding range measurements at the time
  $t_i$ and $t_j$. The optimization step estimates new velocities
  $\bx=(v\;\omega)^T$ under the current set of correspondences $\mathcal
  C$, using \gls{irls}. In
  \eqref{eq:cost} $\rho$ denotes the Huber robust estimator.
  
  Our algorithm is an instance of \gls{icp} since it proceeds by
  alternating between the data association and optimization 
  steps. In the data association step, the corresponding endpoints are found 
  using the nearest neighbor strategy based on current velocity estimates. In the optimization step, the
  velocities are refined from the new correspondences found by the data association.

  \subsection{Velocity Based De-skewing}
  \label{subsec:deskewing}
  Whereas our approach can be applied to more complex kinematics, for
  the sake of simplicity we detail the common case of a unicycle mobile
  base. Our goal is to estimate the location $\bT(\bx, t_i)= \left
  ( \begin{matrix} x_i & y_i & \theta_i \end{matrix} \right )^T$ of the
  mobile base at time $t_i$, assuming it starts from the origin and
  progresses with constant translation and angular velocities over
  sampling time $\bx=(v \hspace{0.2 cm} \omega)^T$.
  After a time $t_i$, the robot would have traveled for a distance
  $l_i=v\cdot t_i$ and its angle is $\theta_i = \omega \cdot t_i$.
  The base will move along an arc of radius $R$, such that:
  \begin{eqnarray}
  \label{eq:curveture}
       R &=& \frac{v_i}{\omega_i} =\frac{l_i}{\theta_i} . 
  \end{eqnarray}
  From this consideration, we can easily compute $\bT$ as
  \begin{eqnarray}     
  \bT(\bx,t_i) = 
  \left(
  \begin{array}{c}
  R  \sin{\theta_i}\\
  R (1-\cos{\theta_i} )
  \end{array}
  \right) =
  \underbrace{v \cdot t_i}_{l_i} \left(
  \begin{array}{c}
  \frac{\sin{\theta_i}}{\theta_i}\\
  \frac{1-\cos{\theta_i}}{\theta_i}
  \end{array}
  \right)
  \label{eq:T}
  \end{eqnarray}

  Assuming the sensor (\gls{lidar}) is located at the center of the
  mobile base, if at time $t_i$ the beam has a relative angle $\alpha_i$
  and reports a range measurement $r_i$, we can straightforwardly find
  the 2D laser endpoint $\bp_i$ as follows:
  \begin{eqnarray}
  \bp_i(\bx,t_i) &=& \bR(\theta_i + \alpha_i) {\begin{pmatrix} r_i \\ 0 \end{pmatrix}}+ \begin{pmatrix} x_i \\ y_i \end{pmatrix}. 
  \label{eq:point}
  \end{eqnarray}
  Here $\bR(\theta_i + \alpha_i)\in SO(2)$ is 2D rotation matrix of
  $\theta_i + \alpha_i$. 
  Applying this process to all the measurements and
  mapping all reconstructed points back to the pose where the
  acquisition of the current scan was started, results in the
  desired de-skew operation.
  
  \subsection{Error Metric}
  \label{error}
  To evaluate a plane-to-plane distance the algorithm needs to compute
  the normal vectors, which are based on the endpoints. Since the
  position of the endpoints is a function of the estimated velocities
  $\bx$, also the normal vectors are. Hence the algorithm has to recompute both
  endpoints and normals at each iteration. Furthermore, when subsequent
  endpoints are too close, the noise affecting the range might result in
  an unstable normal vector, which hinders the error metric. To lessen
  this effect, before each iteration, we regularize the scan to retain
  only temporally subsequent measurements that are sufficiently far from
  each other to ensure a stable normal. In our experiments, we set this
  threshold to 0.15 m. Similarly, if there is a large distance gap
  between two subsequent endpoints ($>$0.4 m), likely, the surface is
  not continuous at that point, hence we drop those measurements too.
  At the end of this regularization step, we end up with a set of
  reasonably stable measurements we can use for the remainder of the
  computation. For each temporally subsequent pair of endpoints, we
  compute a planar patch $\bm_i=\left<\bc_i, \bn_i, t_i\right>$, characterized by a
  center $\bc_i$, a normal vector $\bn_i$, and a timestamp $t_i$, such that:
  \begin{eqnarray}
  \label{eq:midpoint}
  \bc_i &=& \frac{1}{2} \left ( \bp_{i+k}+\bp_{i} \right )  , \\
  \nonumber \\\bn_i &=& \left(
  \begin{array}{cc}
    \hspace{0.1 cm} 0 & 1\\
    -1 & 0
  \end{array}
  \right)
  \frac{\bp_{i+k}-\bp_{i}}
       {\lVert \bp_{i+k}-\bp_{i} \rVert},  \\\nonumber\\
  t_i &=&  \frac{1}{2} \left ( t_{i+k}+t_{i} \right ).
  \end{eqnarray}
  Here, the index $k>0$ accounts for endpoints suppressed during
  regularization. If two planar patches $\bm_i$ and $\bm_j$ corresponds 
  to the same portion of the environment, we can calculate an error
  vector $\be(\bx,t_i,t_j)$ accounting for both their differences in position and
  orientation. The error vector is a function of the velocities, since the patches $\bm_i$ and $\bm_j$
  are computed based on the endpoints. The latter is related to the velocities by~\eqref{eq:point}.

  Let $\be(\bx,t_i,t_j) \in \RR^3$ be the error vector, whose components are defined as follows:
  \begin{eqnarray}
  \be(\bx,t_i,t_j) &=& 
  \left( 
  \begin{array}{cc}
  \frac{1}{2} (\bc_i-\bc_j)^T (\bn_i+\bn_j)\\
  \bn_j-\bn_i
  \end{array}
  \right).
  \label{eq:error}
  \end{eqnarray}
  Here the first dimension accounts for the distance between two corresponding planar
  patches $\bm_i$ and $\bm_j$ projected along the average of their normals. The other two
  account for the difference between the normal vectors.
  \eqref{eq:error} is differentiable in the velocities $\bx$, hence we can minimize \eqref{eq:cost} by Iterative Reweighted Least Squares.

  \begin{figure} 
    \centering
    \includegraphics[width = 0.80\columnwidth]{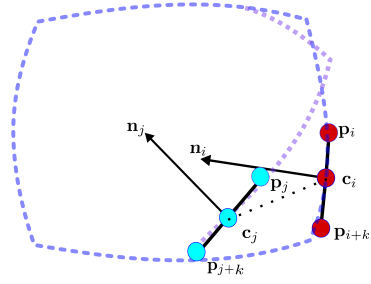}
    \caption{
      Illustration of the planar patches calculation and the correspondence search between two
      matching patches $\bm_i$ and $\bm_j$.}
    \label{fig:skewedScan}
\end{figure}
  \subsection{Data Association}
  \label{subsec:data-association}

  To determine the correspondence we proceed at each iteration by
  de-skewing the sequence of measurements, to get a set of updated
  endpoints $\bp_{i} $. Subsequently, we apply the regularization to
  discard those measurements whose endpoints fall either too close or too
  far from their temporal neighbors. This gives us a set of sequential
  stable measurements we can use to extract the planar patches.
  \figref{fig:skewedScan} demonstarates the data association procedure.

  Once this is done, for each $\bm_i$ we seek for those other patches $\bm_j$ that fulfill all the following criteria:
  \begin{itemize}
      \item their centers are close enough $\lvert \bc_i - \bc_j \rvert<\tau_\mathbf{c}$,
      \item their normals are sufficiently parallel $\bn_i \cdot \bn_j>\tau_\mathbf{n}$,
      \item their timestamp are sufficiently distant $\lvert t_i - t_j \rvert >  \tau_t$.
  \end{itemize}
  Within this set we select as correspondence for $\bm_i$, which is the $\bm_j$ having the smallest projective distance $(\bc_i - \bc_j)^T (\bn_i+\bn_j)$.

\begin{figure}
  \centering
  \includegraphics[width=0.90\columnwidth]{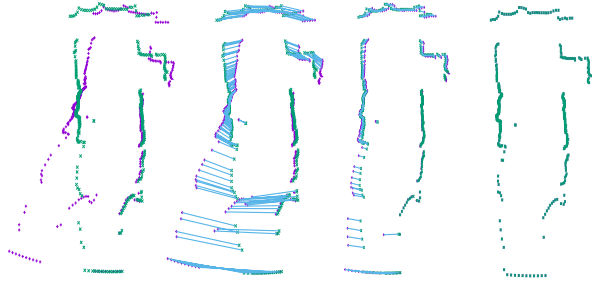}
  \caption{Iterative evolution of the de-skewing process. Purple points represent skewed points,
   green points are the ground truth points, and blue lines represent the data association.
   Initial configuration, instance of full scan acquisition, with RMSE = 0.4 m. 
   After 4 iterations of processing (de-skewing) the raw scan with the proposed approach, with decreased RMSE error to = 0.2 m.
    After 20 iterations, we ended up with a de-skewed scan closer to the ground truth, with RMSE = 0.074 m.}
  \label{fig:scan_evolution}
  \end{figure}

%%%%%%%%%%%%%%%%%%%%%%%%%%%%%%%%%%%%%%%%%%%%%%%%%%%%%%%%%%%%%%%%%%%%%%%%%%%%%%%%
\section{Experimental Validation}
\label{sec:exp}
In this section, we present quantitative evaluations to
demonstrate the accuracy of the velocities estimated by our approach, 
and its effect on the processed data, using only \gls{lidar} measurements. To this extent,
we carried out statistical experiments under changing velocities.

The goal of the first experiment is to compare the velocities of the robot
calculated by our approach with the ground truth applied. As mentioned before
 utilizing the estimated velocities by our approach, renders measurements geometrically consistent
 as seen in \figref{fig:scan_evolution}.
 
 \begin{table*}[h]

  \resizebox{\textwidth}{!}{\begin{tabular}{c||c|c|c|c|c|c} 
  \backslashbox[20mm]{$\omega(rad/s)$}{$v(m/s)$} &-2.000 &-1.000 &-0.500 &0.500 & 1.000 &2.000
    \\ [0.5ex]
    \hline 
     \hline\multirow{3}{*}{-2.000}
    &-1.936$\pm$0.090 & -0.950$\pm$0.068&-0.471$\pm$0.050&0.470$\pm$0.047 &0.962$\pm$0.052&1.910$\pm$0.092\\ 
    &-1.952$\pm$0.081  & -1.933$\pm$0.115&-1.958$\pm$0.066&-1.962$\pm$0.052 &-1.952$\pm$0.070&-1.932$\pm$0.103\\ 
    &$\mathbf{0.090}$$/$0.404& $\mathbf{0.083}$$/$0.399&$\mathbf{0.059}$$/$0.351&$\mathbf{0.061}$$/$0.414 &$\mathbf{0.055}$$/$0.460&$\mathbf{0.081}$$/$0.579\\
    \hline\multirow{3}{*}{-1.000}
    &-1.890$\pm$0.131 & -0.979$\pm$0.031&-0.477$\pm$0.035&0.482$\pm$0.034 &0.946$\pm$0.069&1.897$\pm$0.073\\ 
    &-0.949$\pm$0.069  & -0.986$\pm$0.023&-0.973$\pm$0.037&-0.979$\pm$0.026 &-0.953$\pm$0.059&-0.950$\pm$0.056\\ 
    &$\mathbf{0.067}$$/$0.297& $\mathbf{0.058}$$/$0.308&$\mathbf{0.055}$$/$0.296&$\mathbf{0.049}$$/$0.354 &$\mathbf{0.054}$$/$0.336&$\mathbf{0.062}$$/$0.399\\
    \hline\multirow{3}{*}{-0.500}
    &-1.929$\pm$0.085 & -0.978$\pm$0.063&$-0.479\pm$0.054&0.492$\pm$0.033 &0.935$\pm$0.064&1.906$\pm$0.062\\ 
    &-0.487$\pm$0.020  & -0.493$\pm$0.022&-0.477$\pm$0.035&-0.495$\pm$0.009 &-0.484$\pm$0.026&-0.482$\pm$0.022\\ 
    &$\mathbf{0.040}$$/$0.308& $\mathbf{0.035}$$/$0.345&$\mathbf{0.041}$$/$0.188&$\mathbf{0.043}$$/$0.200 &$\mathbf{0.060}$$/$0.218&$\mathbf{0.084}$$/$0.338\\
    \hline\multirow{3}{*}{0.500}
    &-1.955$\pm$0.024 & -0.954$\pm$0.044&-0.474$\pm$0.071&0.481$\pm$0.048 &0.969$\pm$0.042&1,967$\pm$0.036\\ 
    &0.483$\pm$0.012  & 0.495$\pm$0.018&0.476$\pm$0.045&0.485$\pm$0.026 &0.488$\pm$0.020&0.496$\pm$0.012\\ 
    &$\mathbf{0.119}$$/$0.158& $\mathbf{0.029}$$/$0.140&$\mathbf{0.044}$$/$0.112&$\mathbf{0.052}$$/$0.132 &$\mathbf{0.059}$$/$0.161&$\mathbf{0.159}$$/$0.271\\
    \hline\multirow{3}{*}{1.000}
    &-1.844$\pm$0.130 & -0.976$\pm$0.064&-0.472$\pm$0.057&0.495$\pm$0.028 &0.969$\pm$0.059&1.98$\pm$0.038\\ 
    &0.933$\pm$0.077  & 0.976$\pm$0.053&0.940$\pm$0.053&0.992$\pm$0.015 &0.970$\pm$0.048&0.992$\pm$0.017\\ 
    &$\mathbf{0.063}$$/$0.261& $\mathbf{0.063}$$/$0.225&$\mathbf{0.024}$$/$0.231&$\mathbf{0.055}$$/$0.302 &$\mathbf{0.058}$$/$0.303&$\mathbf{0.039}$$/$0.335\\
    \hline\multirow{3}{*}{2.000}
    &-1.906$\pm$0.116 & -0.941$\pm$0.076&-0.465$\pm$0.081&0.480$\pm$0.065 &0.940$\pm$0.083&1.947$\pm$0.069\\ 
    &1.904$\pm$0.142  & 1.919$\pm$0.120&1.912$\pm$0.104&1.905$\pm$0.137 &1.922$\pm$0.116&1.963$\pm$0.069\\ 
    &$\mathbf{0.074}$$/$0.416& $\mathbf{0.071}$$/$0.368&$\mathbf{0.081}$$/$0.358&$\mathbf{0.075}$$/$0.435 &$\mathbf{0.076}$$/$0.494&$\mathbf{0.091}$$/$0.424\\
    
    \end{tabular}}
   
    \caption{
      Evaluation of the proposed approach using different combinations of translation and angular velocities.
      }
    \label{tab:evaluation}
    \end{table*}
Virtual synthetic scans were created with a skewing effect based on the addressed velocities.
We proceed by de-skewing twice, the first time we compute the ground truth measurement
by setting the applied velocities, and
the second time we compute the de-skewed scan by using our velocity
estimate. The consistency is measured as the distance between
corresponding endpoints in the two scans.
In \tabref{tab:evaluation} each cell contains 3 rows in which: 
The first row is the estimated translational velocity$\bar(v) \pm std (m/s)$,
while the second row is the estimated angular velocity$\bar(\omega) \pm std (rad/s)$, and finally
the third row is the point to point \gls{rmse} $(m)$ for the de-skewed$/$skewed respectively
 with respect to their ground truth counterpart, for each combination of velocities. 
 \begin{figure}
  \centering
  \begin{subfigure}{0.45\textwidth}
      \centering
      \includegraphics[width=\columnwidth]{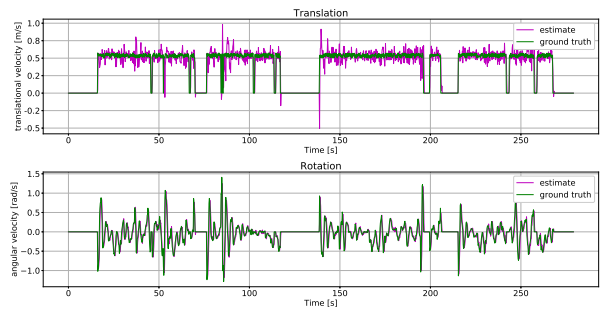} 
      \subcaption{Velocity estimated using synthetic data}
      \label{fig:sim_velocities}
  \end{subfigure}\\
  \begin{subfigure}{0.45\textwidth}
      \centering
      \includegraphics[width=\textwidth]{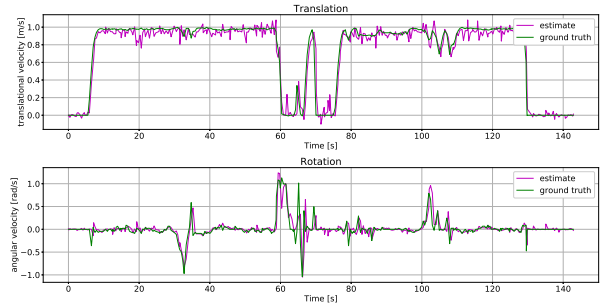}
      \subcaption{Velocity estimated using real data }
      \label{fig:real_velocities}
  \end{subfigure}
 
  \caption{Evolution of estimated and real velocities.}
  \label{fig:velocity-estimates}
\end{figure}

  We repeated this experiment on the real robot and the velocity plots
in~\figref{fig:velocity-estimates} confirm that our system can
effectively recover the robot's velocities, correctly de-skewing the scan, in real indoor scenarios.

  \subsection{De-skewing in a SLAM system}

\begin{table}[!htbp]
  \centering
  \begin{tabular}{*5c}
  \toprule
  \multicolumn{2}{c}{Velocities}&   & \multicolumn{2}{c}{ATE(m)}\\
  \midrule
     $v_\mathrm{max}$ (m/s)& $\omega_\mathrm{max}$(rad/s)&  &  w/o deskewing   &  deskewed\\
  0.5&0.5                        & & 0.329&$\mathbf{0.113}$ \\ 
  0.5&1.0                        & & 0.246&$\mathbf{0.232}$ \\
   1.0&1.0                        & & 1.018&$\mathbf{0.977}$ \\
   1.0&1.5                        & & 1.586&$\mathbf{1.520}$ \\
   2.0&2.0                         & & 11.18&$\mathbf{1.805}$ \\
    \bottomrule
  \end{tabular}
  \caption{SLAM comparison}

  \label{tab:slam}
  \end{table}
  \begin{figure}
    \centering
    \begin{subfigure}{0.22\textwidth}
        \centering
        \includegraphics[width=\textwidth]{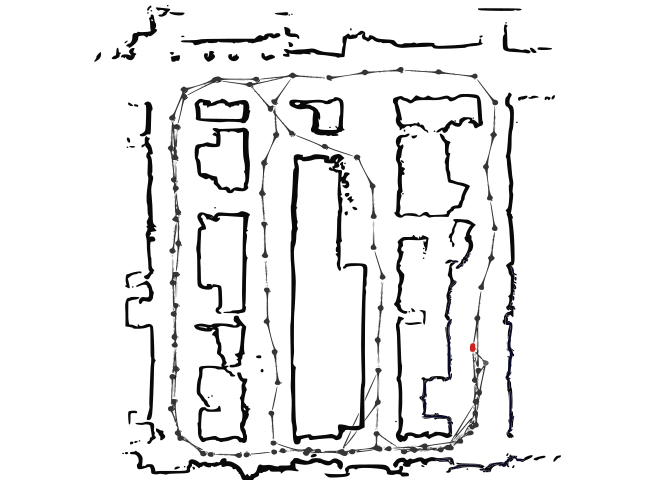}
        \subcaption{}
        \label{fig:gtMap}
    \end{subfigure}
    \begin{subfigure}{0.22\textwidth}
        \centering
        \includegraphics[width=\textwidth]{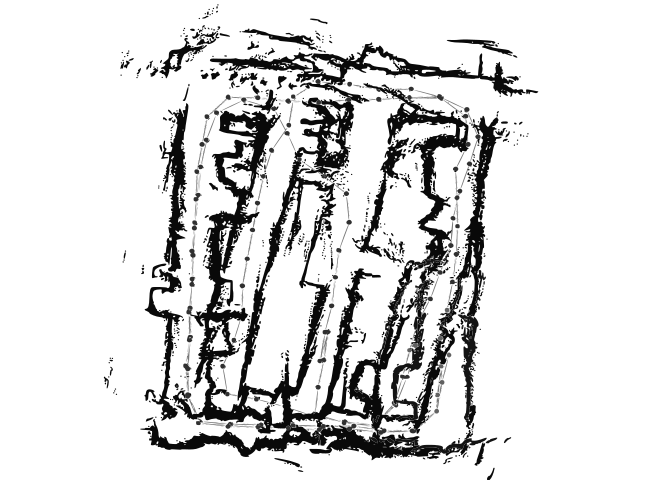}
        \subcaption{}
        \label{fig:skewedMap}
    \end{subfigure}
    \\ 
    \hspace{0.4cm}\begin{subfigure}{0.22\textwidth}
        \centering
        \includegraphics[width=\textwidth]{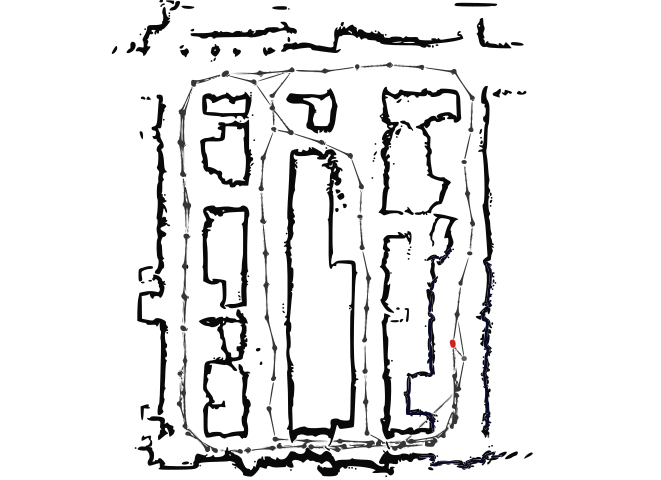}
        \subcaption{}
        \label{fig:desMap}
    \end{subfigure}
    \hspace{0.4cm} \begin{subfigure}{0.2\textwidth}
        \centering
        \includegraphics[width=0.85\columnwidth]{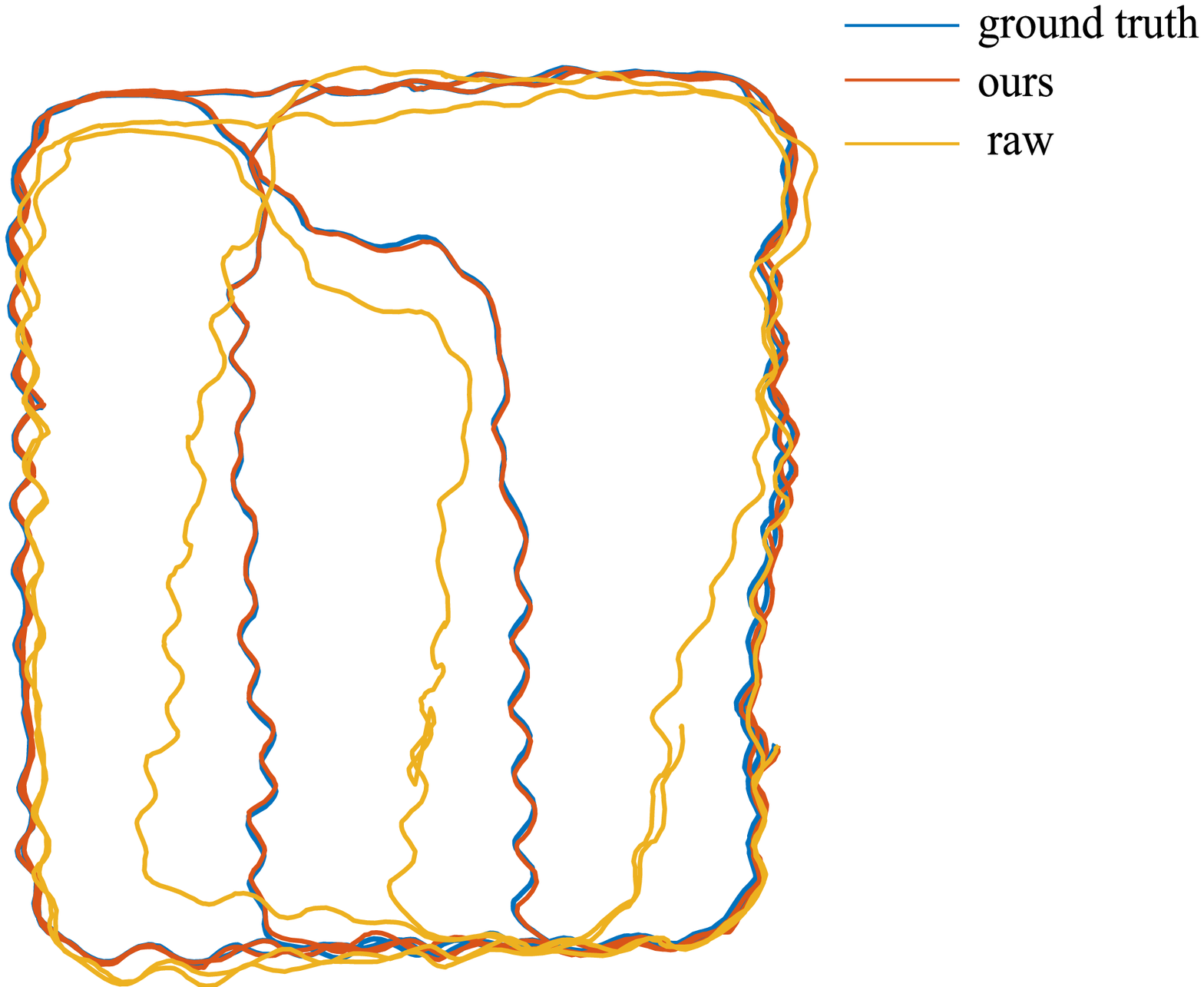}
        \subcaption{}
        \label{fig:trajectories}
    \end{subfigure}
    \caption{SLAM Results. (a) Ground truth map (reference). (b) Map constructed using raw scans acquired directly from the \gls{lidar}. (c) Map constructed using processed \gls{lidar} measurements de-skewed by our proposed approach.
    (d) Estimated trajectories from SLAM, produced by scan matching between consecutive measurements.}
    \label{fig:SLAM}
  \end{figure}
In our final experiment, we used our de-skewing mechanism to
pre-process the input of our 2D \gls{lidar} \gls{slam} system
\cite{colosi2020pnpslam}. We run \gls{slam} on three different
inputs: the raw (skewed) scans, the ones de-skewed by using the proposed
approach, and finally the data de-skewed by using the ground truth.
We simulated different exploration runs of the same environment while
changing the velocity bounds of the robot. \figref{fig:SLAM}
illustrates the three different maps obtained.

Furthermore, we measure the Absolute Trajectory Error (ATE) between the
ground truth trajectory computed by the simulator and the ones
estimated by SLAM  using the two types of scans. ATE measures the
distance between corresponding points of the trajectories, after
computing a transformation that makes them as close as possible. For
trajectory registration, we used the Horn method~\cite{horn1988closed},
while the corresponding poses are associated based on the timestamps.
 The results are summarized in~\tabref{tab:slam}. Consistently, the
trajectories obtained by using de-skewed data are substantially closer
to the ground truth compared to their skewed counterpart as shown in \figref{fig:trajectories}. This is confirmed by the
maps generated based on the trajectories and illustrated in \figref{fig:SLAM}.
These results are consistent and confirm the importance of the de-skewer since these measurements might induce systematic
unrecoverable errors in the registration process.

%%%%%%%%%%%%%%%%%%%%%%%%%%%%%%%%%%%%%%%%%%%%%%%%%%%%%%%%%%%%%%%%%%%%%%%%%%%%%%%%
\section{Conclusion}
\label{sec:conclusion}
In this paper, we presented a simple and effective planar \gls{lidar}
de-skewing mechanism based on plane-to-plane registration
criteria. To the best of our knowledge, this is the only de-skewing
pipeline that does not rely on additional sensors (i.e. IMU, wheel
encoders), targeting specifically slow and inexpensive
\gls{lidar}s, and enhancing the
quality of the scan. For future work, we are planning to integrate the proposed
approach in more complex systems, while generalizing the approach for 3D data. 
%%%%%%%%%%%%%%%%%%%%%%%%%%%%%%%%%%%%%%%%%%%%%%%%%%%%%%%%%%%%%%%%%%%%%%%%%%%%%%%%

% \bibliographystyle{unsrt}

\bibliographystyle{splncs04}
\bibliography{robots}

\end{document}